\documentclass[conference]{IEEEtran}

\usepackage{cite}
\usepackage{amsmath}
\usepackage[utf8]{inputenc}
\usepackage[english]{babel}
\usepackage{graphicx}
\usepackage{algorithmic}
\usepackage{array}
\usepackage{url}
\usepackage{csquotes}

\begin{document}
%
\title{From Imitation to Prediction, Data Compression vs Recurrent Neural Networks for Natural Language Processing}

\author{\IEEEauthorblockN{Juan Andrés Laura\IEEEauthorrefmark{1},
Gabriel Masi\IEEEauthorrefmark{2} and Luis Argerich\IEEEauthorrefmark{3}}
\IEEEauthorblockA{Departemento de Computación,
Facultad de Ingeniería\\
Universidad de Buenos Aires\\
Email: \IEEEauthorrefmark{1}jandreslaura@gmail.com,
\IEEEauthorrefmark{2}masigabriel@gmail.com,
\IEEEauthorrefmark{3}largerich@fi.uba.ar}}


\maketitle

\begin{abstract}
In recent studies [1][13][12] Recurrent Neural Networks were used for generative processes and their surprising performance can be explained by their ability to create good predictions. In addition, data compression is also based on predictions. What the problem comes down to is whether a data compressor could be used to perform as well as recurrent neural networks in natural language processing tasks. If this is possible, then the problem comes down to determining if a compression algorithm is even more intelligent than a neural network in specific tasks related to human language. In our journey we discovered what we think is the fundamental difference between a Data Compression Algorithm and a Recurrent Neural Network.
\end{abstract}

\section{Introduction}
One of the most interesting goals of Artificial Intelligence is the simulation of different human creative processes like speech recognition, sentiment analysis, image recognition, automatic text generation, etc. In order to achieve such goals, a program should be able to create a model that reflects how humans think about these problems.

Researchers think that Recurrent Neural Networks (RNN) are capable of understanding the way some tasks are done such as music composition, writing of texts, etc. Moreover, RNNs can be trained for sequence generation by processing real data sequences one step at a time and predicting what comes next [1][13].

Compression algorithms are also capable of understanding and representing different sequences and that is why the compression of a string could be achieved. However, a compression algorithm might be used not only to compress a string but also to do non-conventional tasks in the same way as neural nets (e.g. a compression algorithm could be used for clustering [11], sequence generation or music composition).

Both neural networks and data compressors have something in common: they should be able to learn from the input data to do the tasks for which they are designed. In this way, we could argue that a data compressor can be used to generate sequences or a neural network can be used to compress data. In consequence, if we use the best data compressor to generate sequences then the results obtained should be better that the ones obtained by a neural network but if this is not true then the neural network should compress better than the state of the art in data compression.

Our hypothesis is that, if compression is based on learning from the input data set, then the best compressor for a given data set should be able to compete with other algorithms in natural language processing tasks. In the present work, we will analyze this hypothesis for two given scenarios: sentiment analysis and automatic text generation.

\section{Data Compression as an Artificial Intelligence Field}

For many authors there is a very strong relationship between Data Compression and Artificial Intelligence [8][9]. Data Compression is about making good predictions which is also the goal of Machine Learning, a field of Artificial Intelligence. 

We can say that data compression involves two steps: modeling and coding. Coding is a solved problem using arithmetic compression. The difficult task is modeling. In modeling the goal is to build a description of the data using the most compact representation; this is again directly related to Artificial Intelligence. Using the Minimal Description Length principle[10] the efficiency of a good Machine Learning algorithm can be measured in terms of how good is is to compress the training data plus the size of the model itself.

If we have a file containing the digits of $\pi$ we can compress the file with a very short program able to generate those digits, gigabytes of information can be compressed into a few thousand bytes, the problem is having a program capable of understanding that our input file contains the digits of $\pi$. We can they say that, in order to achieve the best compression level, the program should be able to always find the most compact model to represent the data and that is clearly an indication of intelligence, perhaps even of General Artificial Intelligence. 

\section{RNNs for Data Compression}

Recurrent Neural Networks and in particular LSTMs were used for predictive tasks [7] and for Data Compression [14]. While the LSTMs were brilliant in their text[13], music[12] and image generation[18] tasks they were never able to defeat the state of the art algorithms in Data Compression[14]. 

This might indicate that there is a fundamental difference between Data Compression and Generative Processes and between Data Compression Algorithms and Recurrent Neural Networs.After experiments we will show that there's indeed a fundamental difference that explains why a RNN can be the state of the art in a generative process but not in Data Compression.

\section{Sentiment Analysis}

\subsection{A Qualitative Approach}

The Sentiment of people can be determined according to what they write in many social networks such as Facebook, Twitter, etc.. It looks like an easy task for humans. However, it could be not so easy for a computer to automatically determine the sentiment behind a piece of writing.

The task of guessing the sentiment of text using a computer is known as sentiment analysis and one of the most popular approaches for this task is to use neural networks. In fact, Stanford University created a powerful neural network for sentiment analysis [3] which is used to predict the sentiment of movie reviews taking into account not only the words in isolation but also the order in which they appear. In our first experiment, the Stanford neural network and a PAQ compressor [2] will be used for doing sentiment analysis of movie reviews in order to determine whether a user likes or not a given movie. After that, results obtained will be compared. Both algorithms will use a public data set for movie reviews [17].

It is important to understand how sentiment analysis could be done with a data compressor. We start introducing the concept of using Data Compression to compute the distance between two strings using the \textit{Normalized Compression Distance} [16].

$$
NCD(x,y) = \frac{C(xy)-\min\{C(x),C(y)\}}{\max\{C(x),C(y)\}}
$$

Where $C(x)$ is the size of applying the best possible compressor to $x$ and $C(xy)$ is the size of applying the best possible compressor to the concatenation of $x$ and $y$. 

The NCD is an approximation to the Kolmogorov distance between two strings using a Compression Algorithm to approximate the complexity of a string because the Kolmogorov Complexity is uncomputable. 

The principle behind the NCD is simple: when we concatenate string $y$ after $x$ then if $y$ is very similar to $x$ we should be able to compress it a lot because the information in $x$ contains everything we need to describe $y$. An observation is that $C(xx)$ should be equal, with minimal overhead difference to $C(x)$ because the Kolmogorov complexity of a string concatenated to itself is equal to the Kolmogorov complexity of the string. 

As introduced, a data compressor performs well when it is capable of understanding the data set that will be compressed. This understanding often grows when the data set becomes bigger and in consequence compression rate improves. However, it is not true when future data (i.e. data that has not been compressed yet) has no relation with already compressed data because the more similar the information it is the better compression rate it is. 

Let $C(X_1,X_2...X_n)$ be a compression algorithm that compresses a set of n files denoted by $X_1,X_2...X_n$. Let $P_1,P_2...P_n$ and $N_1,N_2...N_m$ be a set of $n$ positive reviews and $m$ negative reviews respectively. Then, a review $R$ can be predicted positive or negative using the following inequality:

$$C(P_1,...P_n,R)-C(P_1,...,P_n) < C(N_1,...,N_m,R)-C(N_1,...,N_m)$$

The formula is a direct derivation from the NCD. When the inequality is not true, we say that a review is predicted negative. 

The order in which files are compressed must be considered. As you could see from the proposed formula, the review $R$ is compressed last. 

Some people may ask why this inequality works to predict whether a review is positive or negative. So it is important to understand this inequality. Suppose that the review $R$ is a positive review but we want a compressor to predict whether it is positive or negative. If $R$ is compressed after a set of positive reviews then the compression rate should be better than the one obtained if $R$ is compressed after a set of negative reviews because the review $R$ has more related information with the set of positive reviews and in consequence should be compressed better. Interestingly, both the positive and negative set could have different sizes and that is why it is important to subtract the compressed file size of both sets in the inequality.

\subsection{Data Set Preparation}

We used the Large Movie Review Dataset [17] which is a popular dataset for doing sentiment analysis, it has been used by Kaggle for Sentiment Analysis competitions.

We describe the quantity of movie reviews used in the following table.

\begin{table}[!hbt]
\center{
\begin{tabular}{|l|l|l|}
\hline
	 & Positive & Negative\\
\hline
	Total & 12491 & 12499\\
\hline
	Training & 9999 & 9999\\
\hline
	Test & 2492 & 2500\\
\hline
\end{tabular}
}
\end{table}

\subsection{PAQ for Sentiment Analysis}

The idea of using Data Compression for Sentiment Analysis is not new, it has been already proposed in [5] but the authors did not use PAQ.

We chose PAQ [2] because at the time of this writing it was the best Data Compression algorithm in several benchmarks. The code for PAQ is available and that was important to be able to run the experiments. For the Sentiment Analysis task we used PAQ to compress the positive train set and the negative train set storing PAQ's data model for each set. Then we compressed each test review after loading the positive and negative models comparing the size to decide if the review was positive or negative. 

\subsection{Experiment Results}

	In this section, the results obtained are explained giving a comparison between the data compressor and the Stanford’s Neural Network for Sentiment Analysis. 
	
The following table shows the results obtained

\begin{table}[!hbt]
\center{
\begin{tabular}{|l|l|l|l|}\hline
&\textbf{Correct}&\textbf{Incorrect}&\textbf{Inconcluse} \\ \hline
PAQ & 77.20\% & 18.41\% & 	4.39\% \\ \hline
RNN & 70.93\% & 23.60\% & 5.47\% \\ \hline
\end{tabular}
}
\end{table}

As you could see from the previous table, 77.20\% of movie reviews were correctly classified by the PAQ Compressor whereas 70.93\% were well classificated by the Stanford’s Neural Network. 

There are two main points to highlight according to the result obtained:
\begin{enumerate}
\item Sentiment Analysis could be achieved with a PAQ compression algorithm with high accuracy ratio.
\item In this particular case, a higher precision can be achieved using PAQ rather than the Stanford Neural Network for Sentiment Analysis.
\end{enumerate}

We observed that PAQ was very accurate to determine whether a review was positive or negative, the missclassifications were always difficult reviews and in some particular cases the compressor outdid the human label, for example consider the following review:
\newline
\newline
\textit{“The piano part was so simple it could have been picked out with one hand while the player whacked away at the gong with the other. This is one of the most bewilderedly trance­state inducing bad movies of the year so far for me.”}
\newline
\newline
This review was labeled positive but PAQ correctly predicted it as negative, since the review is misslabeled it counted as a miss in the automated test.

\section{Automatic Text Generation}

This module’s goal is to generate automatic text with a PAQ series compressor and compare it with RNN’s results, using specifics metrics and scenarios.

The ability of good compressors when making predictions is more than evident. It just requires an entry text (training set) to be compressed. At compression time, the future symbols will get a probability of occurrence: The greater the probability, the better compression rate for success cases of that prediction, on the other hand, each failure case will take a penalty. At the end of this process, a probability distribution will be associated with that entry data.

As a result of that probabilistic model, it could be possible to simulate new samples, in other words, generate automatic text.

\subsection{Data Model}

PAQ series compressors use arithmetic coding [2], it encodes symbols assigned to a probability distribution. This probability lies in the interval [0,1) and when it comes to arithmetic coding, there are only two possible symbols: 0 and 1. Moreover, this compressor uses contexts, a main part of compression algorithms. They are built from the previous history and could be accessed to make predictions, for example, the last ten symbols can be used to compute the prediction of the eleventh. 

\begin{figure}[htp]
\centering
\includegraphics[scale=0.50]{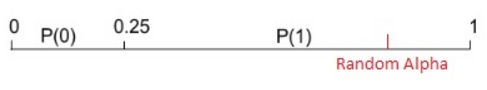}
\caption{we can see that PAQ splits the [0,1) interval giving $1/4$ of probability to the bit 0 and $3/4$ of probability to the bit 1. When a random number is sampled in this interval it is likely that PAQ will generate a 1 bit. After 8 bits are generated we have a character. Each bit generated is used as context but is not used to update the PAQ models because PAQ should not learn from the text it is randomly generating. PAQ learns from the training test and then generates random text using that model.}
\label{fig1}
\end{figure}

PAQ uses an ensamble of several different models to compute how likely a bit 1 or 0 is next. Some of these models are based in the previous $n$ characters of $m$ bits of seen text, other models use whole words as contexts, etc. 
In order to weight the prediction performed by each model, a neural network is used to determine the weight of each model [15]:

$$P(1|c) = \sum_{i=1}^{n} P_i(1|c) W_i$$

Where $P(1|c)$ is the probability of the bit 1 with context "c", $P_i(1|c)$ is the probability of a bit 1 in context "c" for model $i$ and $W_i$ is the weight of model $i$

In addition, each model adjusts their predictions based on the new information. When compressing, our input text is processed bit by bit. On every bit, the compressor updates the context of each model and adjusts the weights of the neural network. 

Generally, as you compress more information, the predictions will be better.

\subsection{Text Generation}

When data set compression is over, PAQ is ready to generate automatic text.

A random number in the $[0,1)$ interval is sampled and transformed into a bit zero or one using Inverse Transform Sampling. In other words, if the random number falls within the probability range of symbol 1, bit 1 will be generated, otherwise, bit 0. 

Once that bit is generated, it will be compressed to reset every context for the following prediction. 

What we want to achieve here is updating models in a way that if you get the same context in two different samples, probabilities will be the same, if not, this could compute and propagate errors. Seeing that, it was necessary to turn off the training process and the weight adjustment of each model in generation time. This was also possible because the source code for PAQ is available.

We observed that granting too much freedom to our compressor could result in a large accumulation of bad predictions that led to poor text generation. Therefore, it is proposed to make the text generation more conservative adding a parameter called “temperature” reducing the possible range of the random number.

On maximum temperature, the random number will be generated in the interval [0,1), giving the compressor maximum degree of freedom to make errors, whereas when the temperature parameter turns minimum, the “random” number will always be 0.5, removing the compressor the degree of freedom to commit errors (in this scenario, the symbol with greater probability will always be generated). 

\begin{figure}[htp]
\centering
\includegraphics[scale=0.20]{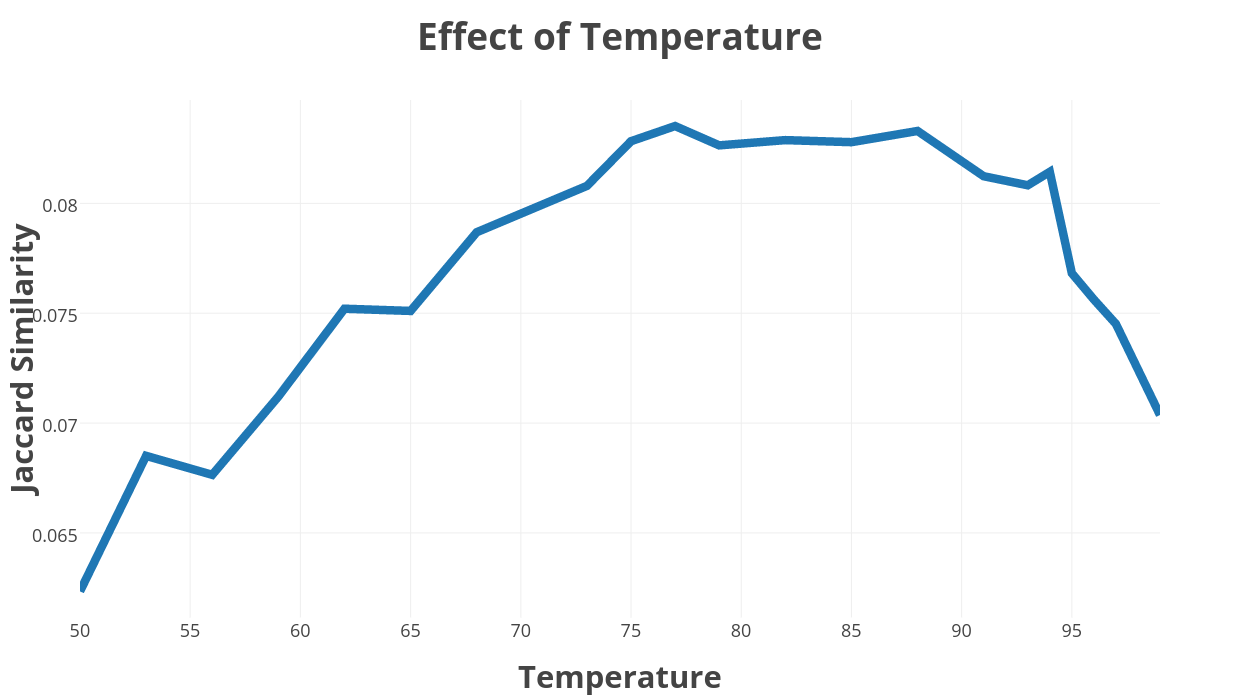}
\caption{Effect of Temperature in the Jaccard Similarity, very high temperatures produce text that is not so similar to the training test, temperatures that are too low aren't also optima, the best value is usually an intermediate temperature.}
\label{}
\end{figure}

When temperature is around 0.5 the results are very legible even if they are not as similar as the original text using the proposed metrics. This can be seen in the following fragment of randomly generated Harry Potter.
\newline

\begin{displayquote}
"What happened?"  said Harry, and she was standing at him.  "He is short, and continued to take the shallows, and the three before he did something to happen again.  Harry could hear him.  He was shaking his head, and then to the castle, and the golden thread broke; he should have been a back at him, and the common room, and as he should have to the good one that had been conjured her that the top of his wand too before he said and the looking at him, and he was shaking his head and the many of the giants who would be hot and leafy, its flower beds turned into the song, and said, "I can took the goblet and sniffed it.  He saw another long for him."
\newline
\end{displayquote}

\subsection{Metrics}

A simple transformation is applied to each text in order to compute metrics.

It consists in counting the number of occurrences of each n-gram in the input (i.e. every time a n-gram ”WXYZ” is detected, it increases its number of occurrences)

Then three different metrics were considered:

\subsubsection{Pearson's Chi-Squared}

How likely it is that any observed difference between the sets arose by chance.

The chi-square is computed using the following formula:

$$
\mathcal{X}^2 = \sum_{i=1}^{n} \frac{(O_i - E_i)^2}{E_i}
$$

Where $O_i$ is the observed ith value and $E_i$ is the expected ith value.

A value of 0 means equality.

\subsubsection{Total Variation}

Each n-gram’s observed frequency can be denoted like a probability if it is divided by the sum of all frequencies, P(i) on the real text and Q(i) on the generated one. 
Total variation distance can be computed according to the following formula:

$$
\delta(P,Q) = \frac{1}{2} \sum_{i=1}^{n} |P_i - Q_i| 
$$

In other words, the total variation distance is the largest possible difference between the probabilities that the two probability distributions can assign to the same event.

\subsubsection{Generalized Jaccard Similarity}
		
It is the size of the intersection divided by the size of the union of the sample sets.

$$
J(G,T) = \frac{G \cap T}{G \cup T}
$$

A value of 1 means both texts are equals.
		
\subsection{Results}

Turning off the training process and the weights adjustment of each model, freezes the compressor’s global context on the last part of the training set. As a consequence of this event, the last piece of the entry text will be considered as a “big seed”.

For example, The King James Version of the Holy Bible includes an index at the end of the text, a bad seed for text generation. If we generate random text after compressing the Bible and its index we get:
\newline
\begin{displayquote}
55And if meat is
   broken behold I will love for the foresaid shall appear, and heard
   anguish, and height coming in the face as a brightness is for God shall
   give thee angels to come fruit.
\newline
\newline
   56But whoso shall admonish them were dim born also for the gift
   before God out the least was in the Spirit into the company
\newline
\newline
   [67Blessed shall be loosed in heaven.)
\end{displayquote}

We noticed this when we compared different segments of each input file against each other, we observed that in some files the last segment was significantly different than the rest of the text. If we remove the index at the end of the file and ask PAQ to generate random text after compressing the Bible we get the following:
\newline
\begin{displayquote}
12The flesh which worship him, he of our Lord Jesus Christ be with you
   most holy faith, Lord, Let not the blood of fire burning our habitation
   of merciful, and over the whole of life with mine own righteousness,
   shall increased their goods to forgive us our out of the city in the
   sight of the kings of the wise, and the last, and these in the
   temple of the blind.
\newline
\newline
  13For which the like unto the souls to the saints salvation, I saw in the
   place which when they that be of the bridegroom, and holy partly, and
   as of the temple of men, so we say a shame for a worshipped his face: I
   will come from his place, declaring into the glory to the behold a
   good; and loosed.
\newline   
\end{displayquote}

The difference is remarkable. It was very interesting to notice that for the RNN the index at the end of the bible did not result in a noticeable difference for the generated text. This was the first hint that the compressor and the RNN were proceeding in different ways.

\begin{figure}[htp]
\centering
\includegraphics[scale=0.20]{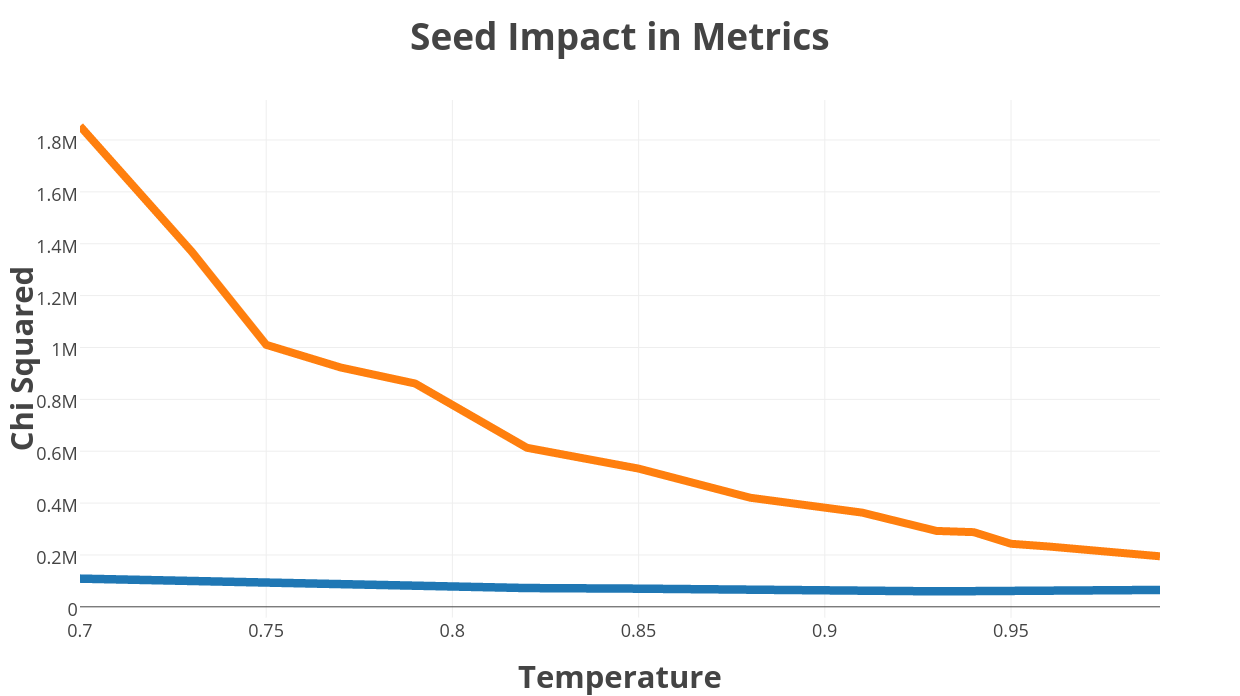}
\caption{The effect of the chosen seed in the Chi Squared metric. In Orange the metric variation by temperature using a random seed. In Blue the same metric with a chosen seed.}
\label{}
\end{figure}

In some cases the compressor generated text that was surprisingly well written. This is an example of random text generated by PAQ8L after compressing "Harry Potter"
\newline
\begin{displayquote}
CHAPTER THIRTY-SEVEN - THE GOBLET OF LORD VOLDEMORT OF THE FIREBOLT MARE!"
\newline
\newline
Harry looked around.  Harry knew exactly who lopsided, looking out parents. They had happened on satin' keep his tables."
\newline
\newline
Dumbledore stopped their way down in days and after her winged around him.
\newline
\newline
He was like working, his eyes.  He doing you were draped in fear of them to study of your families to kill, that the beetle, he time.  Karkaroff looked like this.  It was less frightening you.
\newline
\newline
"Sight what's Fred cauldron bottle to wish you reckon?  Binding him to with his head was handle."
Once and ask Harry where commands and this thought you were rolling one stationed to do.  The stone.  Harry said, battered.
\newline
\newline
"The you," said Ron, and Harry in the doorway. Come whatever Hagrid was looking from understood page.  "So, hardly to you," said Fred, no in the morning.  "They're not enough, we'll to have all through her explain, and the others had relicious importance," said Dumbledore, he wouldn't say anything."
\newline
\end{displayquote}

While the text may not make sense it certainly follows the style, syntax and writing conventions of the training text.

Analyzers based on words like the Stanford Analyzer tend to have difficulties when the review contains a lot of uncommon words. It was surprising to find that PAQ was able to correctly predict these reviews.

Consider the following review:
\newline

\begin{displayquote}
”The author sets out on a "journey of discovery" of his "roots" in the southern tobacco industry because he believes that the (completely and deservedly forgotten) movie "Bright Leaf" is about an ancestor of his. Its not, and he in fact discovers nothing of even mild interest in this absolutely silly and self-indulgent glorified home movie, suitable for screening at (the director's) drunken family reunions but certainly not for commercial - or even non-commercial release. A good reminder of why most independent films are not picked up by major studios - because they are boring, irrelevant and of no interest to anyone but the director and his/her immediate circles. Avoid at all costs!”
\newline
\end{displayquote}

This was classified as positive by the Stanford Analyzer, probably because of words such as "interest, suitable, family, commercial, good, picked", the Compressor however was able to read the real sentiment of the review and predicted a negative label. In cases like this the Compressor shows the ability to truly understand data.

\subsection{Metric Results}

We show the result of bot PAQ and a RNN for text generation using the mentioned metrics to evaluate how similar the generated text is to the original text used for training.

\begin{table}[!hbt]
\center{
\begin{tabular}{|l|l|l|}\hline
&\textbf{PAQ8L} & \textbf{RNN}\\ \hline
Game of Thrones & 47790 & \textbf{44935} \\ 
Harry Potter & \textbf{46195} & 83011 \\
Paulo Coelho & \textbf{45821} & 86854 \\
Bible & \textbf{47833} & 52898 \\
Poe & 	61945 & \textbf{57022} \\
Shakespeare & \textbf{60585} & 84858 \\
Math Collection & \textbf{84758} & 135798 \\
War and Peace & \textbf{46699} & 47590 \\
Linux Kernel & \textbf{136058} & 175293 \\ \hline
\end{tabular}
}
\caption{Chi Squared Results (lower value is better)}
\end{table}

It can be seen that the compressor got better results for all texts except Poe and Game of Thrones.

\begin{table}[!hbt]
\center{
\begin{tabular}{|l|l|l|}\hline
&\textbf{PAQ8L} & \textbf{RNN}\\ \hline
Game of Thrones & 25.21 & \textbf{24.59} \\
Harry Potter & \textbf{25.58} & 37.40 \\
Paulo Coelho & \textbf{25.15} & 34.80 \\
Bible &  \textbf{25.15} & 25.88 \\
Poe & 30.23 & \textbf{27.88} \\
Shakespeare & \textbf{27.94} & 30.71\\
Math Collection & \textbf{31.05} & 35.85 \\
War and Peace & \textbf{24.63} & 25.07 \\
Linux Kernel & \textbf{44.74} & 45.22 \\ \hline
\end{tabular}
}
\caption{Total Variation (lower value is better)}
\end{table}

The results of this metric were almost identical to the results of the Chi-Squared test.

\begin{table}[!hbt]
\center{
\begin{tabular}{|l|l|l|}\hline
&\textbf{PAQ8L} & \textbf{RNN}\\ \hline
Game of Thrones & 0.06118 & \textbf{0.0638} \\
Harry Potter &  \textbf{0.1095} & 0.0387 \\
Paulo Coelho &  \textbf{0.0825} & 0.0367 \\
Bible &  \textbf{0.1419} & 0.1310 \\
Poe & 	0.0602 & \textbf{0.0605} \\
Shakespeare & 0.0333 & \textbf{0.04016} \\
Math Collection & \textbf{0.21} & 0.1626 \\
War and Peace & \textbf{0.0753} & 0.0689 \\
Linux Kernel & \textbf{0.0738} & 0.0713 \\ \hline
\end{tabular}
}
\caption{Jaccard Similarity (higher is better)}
\end{table}

In the Jaccard similarity results were again good for PAQ except for "Poe", "Shakespeare" and "Game of Thrones", there is a subtle reason why Poe was won by the RNN in all metrics and we'll explain that in our conclusions.

\section{Conclusions}

In the sentiment analysys task we have noticed an improvement using PAQ over a Neural Network. We can argue then that a Data Compression algorithm has the intelligence to understand text up to the point of being able to predict its sentiment with similar or better results than the state of the art in sentiment analysis. In some cases the precision improvement was up to 6\% which is a lot. 

We argue that sentiment analysys is a predictive task, the goal is to predict sentiment based on previously seen samples for both positive and negative sentiment, in this regard a compression algorithm seems to be a better predictor than a RNN.

In the text generation task the use of a right seed is needed for a Data Compression algorithm to be able to generate good text, this was evident in the example we showed about the Bible. This result is consistent with the sentiment analysis result because the seed is acting like the previously seen reviews if the seed is not in sync with the text then the results will not be similar to the original text.

The text generation task showed the critical difference between a Data Compression algorithm and a Recurrent Neural Network and we believe this is the most important result of our work: Data Compression algorithms are \textit{predictors} while Recurren Neural Networks are \textit{imitators}.

The text generated by a RNN looks in general better than the text generated by a Data Compressor but if we only generate one paragraph the Data Compressor is clearly better. The Data Compressor learns from the previously seen text and creates a model that is optimal for predicting what is next, that is why they work so well for Data Compression and that is why they are also very good for Sentiment Analysis or to create a paragraph after seeing training test. 

On the other hand the RNN is a great imitator of what it learned, it can replicate style, syntax and other writing conventions with a surprising level of detail but what the RNN generates is based in the whole text used for training without weighting recent text as more relevant. In this sense we can argue that the RNN is better for random text generation while the Compression algorithm should be better for random text extension or completion.

If we concatenate the text of Romeo \& Juliet after Shapespeare and ask both methods to generate a new paragraph the Data Compressor will create a new paragraph of Romeo and Juliet while the RNN will generate a Shakespeare-like piece of text. \textbf{Data Compressors are better for local predictions and RNNs are better for global predictions.}

This explains why in the text generation process PAQ and the RNN obtained different results for different training tests. PAQ struggled with "Poe" or "Game of Thrones" but was very good with "Coelho" or the Linux Kernel. What really happened was that we measured how predictable each author was!. If the text is very predictable then the best predictor will win, PAQ defeated the RNN by a margin with the Linux Kernel and Paulo Coelho. When the text is not predictable then the ability to imitate in the RNN defeated PAQ. This can be used as a wonderful tool to evaluate the predictability of different authors comparing if the Compressor or the RNN works better to generate similar text. In our experiment we conclude that Coelho is more Predictable than Poe and it makes all the sense in the world!

As our final conclusion we have shown that Data Compression algorithms show rational behaviour and that they they are based in the accurate prediction of what will follow based on what they have learnt recently. RNNs learn a global model from the training data and can then replicate it. That's what we say that Data Compression algorithms are great \textbf{predictors} while Recurrent Neural Networks are great \textbf{imitators}. Depending on which ability is needed one or the other may provide the better results. 

\section{Future Work}

We believe that Data Compression algorithms can be used with a certain degree of optimality for any Natural Language Processing Task were predictions are needed based on recent local context. Completion of text, seed based text generation, sentiment analysis, text clustering are some of the areas where Compressors might play a significant role in the near future.

We have also shown that the difference between a Compressor and a RNN can be used as a way to evaluate the predictability of the writing style of a given text. This might be expended in algorithms that can analyze the level of creativity in text and can be applied to books or movie scripts.

\end{document}